\documentclass[10pt,twocolumn,letterpaper]{article}

\usepackage[pagenumbers]{cvpr} 

\definecolor{cvprblue}{rgb}{0.21,0.49,0.74}
\usepackage[pagebackref,breaklinks,colorlinks,allcolors=cvprblue]{hyperref}
\usepackage{xcolor}
\usepackage{mdframed}
\usepackage{makecell}
\usepackage{listings}

\usepackage{bbm}

\definecolor{lightbg}{RGB}{242, 242, 242}
\definecolor{lighttext}{RGB}{30, 30, 30}
\definecolor{lightkeyword}{RGB}{0, 0, 255}
\definecolor{lightcomment}{RGB}{0, 128, 0}
\definecolor{lightstring}{RGB}{163, 21, 21}
\definecolor{lightnum}{RGB}{153, 0, 153}
\definecolor{lightbuiltin}{RGB}{0, 128, 128}
\definecolor{lightidentifier}{RGB}{0, 0, 0}
\definecolor{lightvariable}{RGB}{0, 0, 0}
\definecolor{lightfunction}{RGB}{128, 0, 128}

\lstdefinestyle{pythonstyle}{
    language=Python,
    backgroundcolor=\color{lightbg},
    basicstyle=\footnotesize\ttfamily\color{lighttext},
    keywordstyle=\color{lightkeyword},
    commentstyle=\color{lightcomment},
    stringstyle=\color{lightstring},
    numberstyle=\color{lightnum},
    identifierstyle=\color{lightidentifier},
    tabsize=2,
    showstringspaces=false,
    breaklines=true,
    frame=tb,
    framesep=4pt,
    numbersep=8pt,
    numberstyle=\tiny\color{gray},
    moredelim=[s][\color{lightbuiltin}]{@}{\ },
    morekeywords={self},
}

\definecolor{codegreen}{rgb}{0,0.6,0}
\definecolor{codegray}{rgb}{0.5,0.5,0.5}
\definecolor{codepurple}{rgb}{0.58,0,0.82}
\definecolor{backcolour}{rgb}{0.95,0.95,0.92}

\lstdefinestyle{mystyle}{
    backgroundcolor=\color{backcolour},   
    commentstyle=\color{codegreen},
    keywordstyle=\color{magenta},
    numberstyle=\tiny\color{codegray},
    stringstyle=\color{codepurple},
    basicstyle=\ttfamily\scriptsize,
    breakatwhitespace=false,         
    breaklines=true,                 
    captionpos=t,                    
    keepspaces=true,                 
    numbers=none, 
    numbersep=5pt,                  
    showspaces=false,                
    showstringspaces=false,
    showtabs=false,                  
    tabsize=2,
    showlines=true
}

\def\paperID{18585} 
\def\confName{CVPR}
\def\confYear{2025}

\title{Is Your Text-to-Image Model Robust to Caption Noise?}

\author{
    Weichen Yu\textsuperscript{1,2}\thanks{This work was done during Weichen Yu’s internship at ByteDance.}, 
    Ziyan Yang\textsuperscript{1}, 
    Shanchuan Lin\textsuperscript{1}, 
    Qi Zhao\textsuperscript{1}, 
    Jianyi Wang\textsuperscript{1}, 
    Liangke Gui\textsuperscript{1}, \\
    Matt Fredrikson\textsuperscript{2}, 
    Lu Jiang\textsuperscript{1}\\[1ex]
    { \textsuperscript{1}ByteDance, }
    { \textsuperscript{2}Carnegie Mellon University}\\
   {\small \textsuperscript{1} \texttt{\{ziyan.yang,peterlin,kevin.zhao,jianyi.wang,liangke.gui,lu.jiang\}@bytedance.com}} \\ 
   {\small \textsuperscript{2} \texttt{\{weichenyu,mfredrik\}@cmu.edu} }
}

\begin{document}
\maketitle
\begin{abstract}
In text-to-image (T2I) generation, a prevalent training technique involves utilizing Vision Language Models (VLMs) for image re-captioning. Even though VLMs are known to exhibit hallucination, generating descriptive content that deviates from the visual reality, the ramifications of such caption hallucinations on T2I generation performance remain under-explored. Through our empirical investigation, we first establish a comprehensive dataset comprising VLM-generated captions, and then systematically analyze how caption hallucination influences generation outcomes. Our findings reveal that \textbf{(1)} the disparities in caption quality persistently impact model outputs during fine-tuning. \textbf{(2)} VLMs confidence scores serve as reliable indicators for detecting and characterizing noise-related patterns in the data distribution. \textbf{(3)} even subtle variations in caption fidelity have significant effects on the quality of learned representations. These findings collectively emphasize the profound impact of caption quality on model performance and highlight the need for more sophisticated robust training algorithm in T2I. In response to these observations, we propose a approach leveraging VLM confidence score to mitigate caption noise, thereby enhancing the robustness of T2I models against hallucination in caption.
\end{abstract}

\section{Introduction}

\begin{figure}[th]
    \begin{center}
    \includegraphics[width=0.85\linewidth]{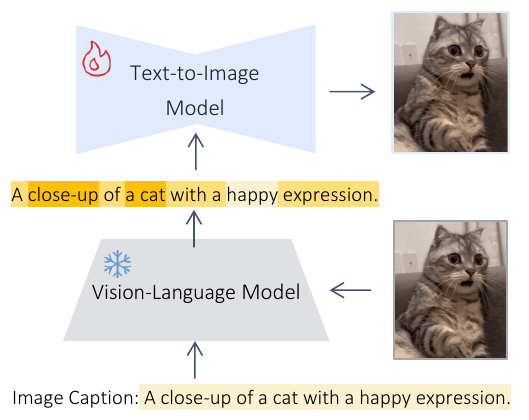}
    \caption{Hallucinated contents exist in image captions. We propose to use pre-trained vision-language models to provide signals in guiding text-to-image models to focus on more reliable terms.}
    \label{fig:lead}
    \end{center}
    \vspace{-2em}
\end{figure}

A text-to-image (T2I) model generates an image based on a natural language description, referred to as a prompt, aiming at aligning the output image with the given description. Recent advancements in T2I generation have showcased impressive capabilities in producing high-fidelity and diverse visual content~\cite{ramesh2022hierarchical, saharia2022photorealistic, rombach2022high, rezende2015variational}. \footnote{Work in progress.}

Enormous training data, specifically image-text pairs, is essential for training or fine-tuning T2I models. There has been a notable shift in the approach to acquiring these text descriptions. Early works~\cite{nichol2021glide} collect image-text pairs by harvesting their structural associations on the web, such as images and accompanying text found in the same context on web pages, or search queries paired with matched images. Since these texts are written by humans, they tend to be concise, often fewer than a dozen words, and focus on describing the most essential elements of the image. This approach remained dominant until the introduction of DALL-E 3~\cite{betker2023improving}, which proposes leveraging large vision-language models (VLMs) to generate long and highly detailed captions, typically more than a hundred words, which describe images with much greater specificity. DALL-E 3 demonstrates the importance of detailed and descriptive captions for enhancing prompt following and handling complex user prompts. Consequently, since DALL-E 3, the use of VLM-generated captions to curate image-text paired data has become a common practice, influencing the development of subsequent T2I models~\cite{chen2024pixart,saharia2022photorealistic,segalis2023picture,esser2024scaling}.

VLMs can make two types of errors: \emph{omission} and \emph{hallucination}.~\emph{Omission} occurs when a concept present in the image is not captured by the VLM, while \emph{hallucination} refers to the VLM generating an object that does not exist in the image or describing an action that is inaccurately represented in the image. Previous studies have suggested that hallucinations can significantly impair the performance of T2I models~\cite{li2023evaluating,gunjal2024detecting}. Relying solely on more advanced VLMs to address this problem is not ideal, as even state-of-the-art VLMs are not free from hallucinations, and can introduce systematic errors in certain aspects as shown in~\cite{tong2024eyes,liu2023mitigating}, such as spatial relationships, quantities, \etc.

On the other hand, hallucinations in captions is a form of label noise. Although our community has developed extensive robust learning methods for mitigating label noise in classification, clustering~\cite{}, little attention has been paid to addressing label noise in text-to-image generation. A very recent study~\cite{na2024label} demonstrates that a simple robust learning approach, label flipping, can effectively improve T2I pretraining. Prior studies primarily address label noise or complete caption errors, which represent extreme cases that rarely occur in practical scenarios.

In this paper, we investigate the subtle role of hallucinations in influencing text-to-image (T2I) generation. First, we quantify the impact of hallucinations on text-to-image generation outcomes. To achieve this, we construct a benchmark dataset of XXX image captions generated by three models: LLaVA-7B-v1.6~\cite{liu2024llava}, Share-Captioner~\cite{chen2023sharegpt4v}, and XX. The first two represent leading open-source captioners,  while the third serves as a weaker baseline for controlled comparisons. We consider continuous training from the same pretrained model to efficiently evaluate performance deviations resulting from the quality of the captioning model. Given that commonly used T2I quality metrics for text alignment, such as CLIP-Score~\cite{radford2021learning}, appear insufficient for evaluating generation quality, we draw inspiration from~\cite{li2023evaluating} and incorporate additional factors such as spatial relationships, quantity, and color into our evaluation. Additionally, we employ linear probing—a technique commonly used in discriminative models—to assess the quality of the learned T2I feature representations.

Through our analysis, we observe that VLMs exhibit patterns in generating hallucinated content. Specifically, the VLM token prediction scores may be used to statistically differentiate hallucinated tokens from normal ones. Building on this insight, we propose an embarrassingly straightforward method to mitigate hallucination by realigning the weights used in computing attention during T2I training. It is worth noting that VLM prediction statistics, which are central to our approach, are typically discarded and overlooked in the existing T2I methods.

Our method demonstrates improved performance in both standard text-to-image metrics (CLIP-Score and FID) and instruction adherence on our benchmark. Linear probing analysis further reveals enhanced quality of intermediate feature representations.

The contribution of this paper are threefold.
 \begin{enumerate}
     \item We conduct an in-depth analysis of the impact of subtle hallucinations in captions on the performance of text-to-image (T2I) generation. Our findings highlight the critical need to address caption noise, underscoring its role in enhancing model robustness.
     \item We identify the utility of vision-language model (VLM) confidence scores in distinguishing hallucinated patterns from clean patterns, providing a simple yet useful indicator for addressing caption hallucination.
     \item We propose a straigthforward approach that leverages VLM confidence scores to dynamically reweight tokens during T2I model training. Experimental results validate the efficacy of this method, demonstrating improvements in model performance and resilience to noisy data.
 \end{enumerate}

\section{Related Work}

\paragraph{Text-to-Image Generation.} Various approaches~\cite{ho2020denoising, goodfellow2014generative, heusel2017gans, rezende2015variational} have been proposed to generate image from natural language. Among those, autoregressive methods~\cite{ramesh2021zero,ding2021cogview,yu2022scaling,kondratyuk2023videopoet} and diffusion models~\cite{rombach2022high,chen2024pixart,ramesh2022hierarchical,song2019generative} have shown promising progress after trained on large-scale internet image-text pairs followed by data filtering steps. To improve text-to-image generative models, previous works mainly focus on the visual side, such as latent diffusion models~\cite{rombach2022high, ramesh2022hierarchical}, transformer based diffusion~\cite{peebles2023scalable, chang2022maskgit, chang2023muse}, or style generations~\cite{sohn2023styledrop, zhao2024uni, ruiz2023dreambooth}. The evaluation of these generative models are based on general metrics such as FID score~\cite{heusel2017gans} for image generation quality and CLIP score ~\cite{hessel2021clipscore} for textual alignment. Our work focuses on the text side of the text-to-image generation. We bring attention to the nuances within image captions and their impact on T2I generation and provide more dimensions of evaluation, which help improve many previously overlooked areas such as hallucinations in generation, robustness in T2I, and \etc. 

\paragraph{Caption Quality in T2I.} 
Recent literature has increasingly emphasized the critical role of caption quality in text-to-image (T2I) generation models, leading to a paradigm shift in dataset preparation methodologies~\cite{segalis2023picture}. Contemporary state-of-the-art T2I architectures have adopted sophisticated vision-language models (VLMs)~\cite{liu2024visual,wang2023cogvlm,bai2023qwen,cheng2023vindlu} for image recaptioning to enhance training data quality. Notably, PixArt-Alpha~\cite{chen2023pixartalpha} and PixArt-Sigma~\cite{chen2024pixart} utilize LLaVA-7B-v1.5~\cite{liu2024visual} and Share-Captioner~\cite{chen2023sharegpt4v} for recaption the training dataset respectively, though the former caption model LLaVA1.5 is identified significant hallucination tendencies in its outputs. Similarly, ~\citep{segalis2023picture} employs PaLI~\cite{chen2022pali}, while Stable Diffusion 3~\cite{esser2024scaling} leveraged CogVLM~\cite{wang2023cogvlm} for caption generation. While prior studies have addressed hallucination detection and mitigation in VLMs~\cite{varshney2023stitch}, these techniques remain unexplored in text-to-image recaptioning tasks.

\paragraph{T2I Noise Robustness.} While noise robustness has been extensively examined within understanding tasks~\cite{natarajan2013learning,jiang2018mentornet}, it has also garnered interest in generative tasks, particularly in the context of Generative Adversarial Networks (GANs)~\cite{kaneko2019label, chrysos2020rocgan,thekumparampil2018robustness,katsumata2024soft,kaneko2020noise,kaneko2021blur}, though less explored in diffusion frameworks~\cite{na2024label,chen2024slight}. Within GAN frameworks, a body of research enhances robustness by estimating clean-label conditional generative distributions~\cite{kaneko2019label, chrysos2020rocgan}. A regularization approach has proven beneficial, particularly by incorporating a permutation regularizer into the adversarial loss function~\cite{thekumparampil2018robustness}. Additionally, \citeauthor{katsumata2024soft} apply curriculum learning to further bolster robustness, while \citeauthor{kaneko2020noise} assume specific noise characteristics, encouraging the noise generator to model noise-specific elements explicitly. Robustness against image noise is further achieved by blending image data with noise~\cite{kaneko2021blur}. However, these studies often operate under assumptions about transition between clean labels and noisy labels, which may not effectively address the complexities introduced by caption noise.

In recent developments within diffusion frameworks, for label noise, \citeauthor{na2024label} interpret noisy-label conditions through a linear combination with clean-label conditions. However, this method requires training a noisy-label classifier and a conditional score model, necessitating a multi-stage training process. For caption noise, minimal noise in diffusion models has been shown to be advantageous~\cite{chen2024slight}.

\section{Robustness to Caption Noise in T2I}

In this section, we first observe that VLMs exhibit patterns in generating hallucinated contents -- we compute three distinct types of confidence scores using VLM, and then investigate the correlation between these scores and hallucinated content. Building on these insights, as illustrated in \cref{fig:pipeline}, we propose a simple method -- to use token reweighting to leverage the computed confidence scores to mitigate the effect of hallucinations in T2I model training and improve robustness to caption noise.

\subsection{Correlation between Bias in VLM and Caption Noise}
\noindent
\textbf{Motivation}. We aim to distinguish hallucinated patterns from clean patterns. Given that current Vision-Language Models (VLMs) are typically trained using either a fixed Language Model (LLM) encoder or with minimal modifications to the LLM encoder,there are inherent biases in the text modality. Specifically, we seek to determine whether hallucinated content in VLM outputs is more dependent on the textual input than on the visual input.

To explore this hypothesis, we begin by constructing a validation subset. Using GPT-4o, we label a sample of 1,000 image-caption pairs, identifying words in the captions that conflict with the content of the images—\ie, hallucinated elements in the captions. Detailed prompt used for the GPT-4o is provided in the supplementary. Subsequently, we employ the VLM to quantify hallucination scores using three distinct confidence scores, providing insights into the dependence of hallucination on text-based biases.

Given an image $I \in \mathcal{R}^{H \times W}$ \footnote{In this paper, we resize image to $256 \times 256$ during training and inference.} and the associated caption $c=[c_0, c_1, \cdots]$, where $c_k$ denotes the $k$-th token, we utilize a VLM, denoted by $\mathcal{P}$, to compute a confidence score $s_k$ for each individual token $c_k$ in the caption. We calculate the confidence score in the following three ways:
Confidence score is calculated only on every text tokens, while the input of the VLM includes both image and caption:
\begin{equation}
    s_k^i = \mathcal{P}(c_k|I,c_{<k})
\end{equation}
Confidence score without image, text only:
\begin{equation} \label{eq:text_only}
    s_k^t = \mathcal{P}(c_k|c_{<k})
\end{equation}
Confidence score difference between with image and without image:
\begin{equation}
    s_k^d = \mathcal{P}(c_k|c_{<k}) - \mathcal{P}(c_k|I,c_{<k})
\end{equation}

\begin{figure}[th]
    \centering
    \includegraphics[width=\linewidth]{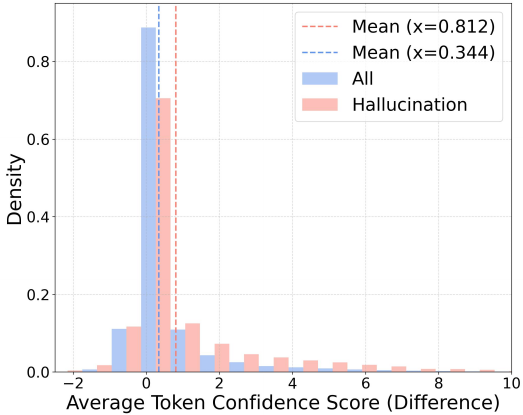}
    \caption{Histogram of token confidence score. The blue bars illustrate the confidence score distribution across all tokens, while the red bars focus specifically on hallucinated tokens. The dashed line denotes the mean confidence score of the corresponding tokens. The differential token confidence scores, obtained by subtracting the text-only score from the text-and-image score. \textbf{The histogram of hallucinated tokens exhibits subtle but distinct variations compared to clean tokens, indicating measurable differences in their distributional characteristics}.
}\label{fig:hist_d}
\end{figure}

As shown in \cref{fig:hist_d}, we observe that there's a distribution shift between the noisy tokens and all the tokens. This observation suggests that the hallucinated content in the caption can be captured by the Vision-Language Model’s (VLM) confidence scores. Specifically, the confidence scores, $s^d$, indicate the VLM's degree of "confidence" difference when generating a given token with image and without image.

\subsection{An Approach to Alleviate Caption Noise in T2I} \label{sec:approach}

\begin{figure*}[th]
    \begin{center}
    \includegraphics[width=0.88\linewidth]{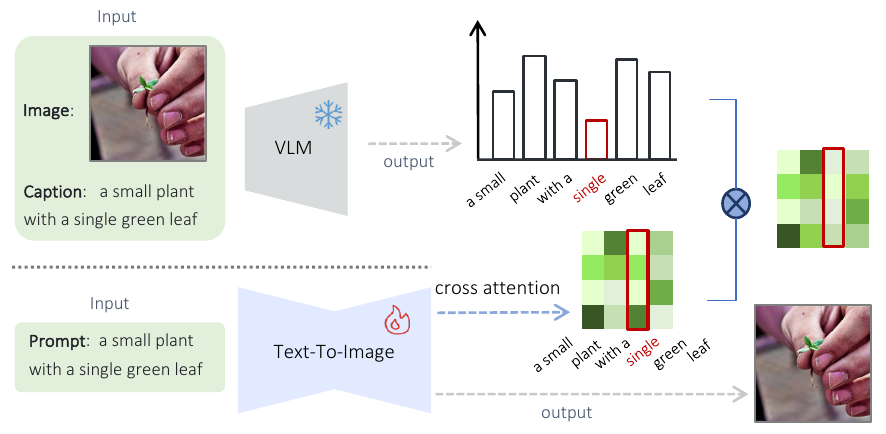}
    \caption{An overview of our proposed method to alleviate caption noise using a pre-trained VLM. We use a pre-trained VLM to compute the confidence score of each token in the caption. And then we use confidence score to reweight each token in T2I training, alleviating the effect of possibly hallucinated contents in captions.}
    \label{fig:pipeline}
    \end{center}
    \vspace{-0.5cm}
\end{figure*}

Based on the observation that VLM can be an indicator, outputing a confidence score for each token in the caption, we leverage this score to downweighting the tokens that have a higher possibility to be noisy.

In the T2I model, there are layers with CrossAttention that uses the text to instruct the image generation. Thus, a straightforward method is to assign different weights to different tokens. We have also tried some other methods, such as drop out the probably `noisy' tokens and concatanate other tokens as filtered caption, but the results are not as ideal.

As in the \cref{fig:pipeline}, first we compute the confidence score of each token in the caption, and then we map the confidence to the tokens used in the pipeline,

\subsubsection{Tokenizer Mapping}
Since the VLMs we use to calculate the confidence score and the T2I model do not use a unified tokenizer, a tokenizer mapping to align the different tokenizers is needed.

Recent advancements in vision-language models (VLMs) and large language models (LLMs) have increasingly adopted distinct, non-unified tokenization schemes, necessitating robust handling of diverse tokenizers within multi-model frameworks. In response, we introduce a tokenizer mapping strategy designed to align tokenized representations across models.

\begin{figure}[th]
    \begin{center}
    \includegraphics[width=\linewidth]{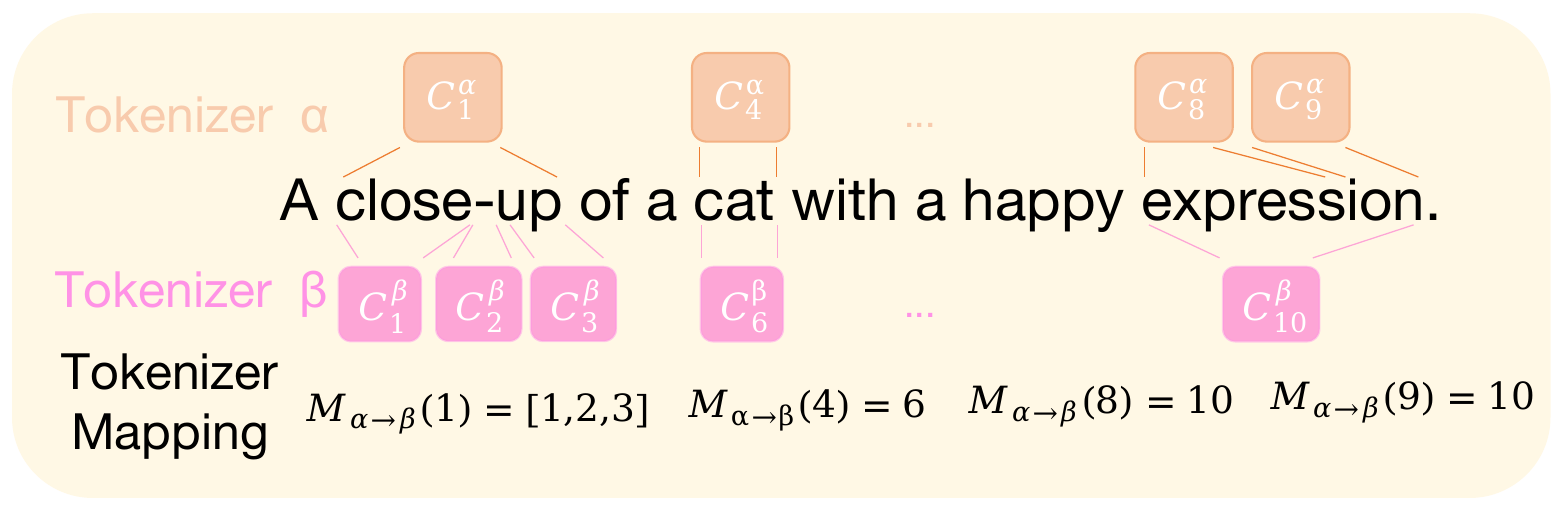}
    \caption{Illustration of tokenizer mapping. Given two distinct tokenizers, this tokenizer mapping allows us to construct a mapping function from one encoder output to another.}
    \label{fig:t_map}
    \end{center}
    \vspace{-0.5cm}
\end{figure}

Suppose we have two distinct tokenizer, $\mathcal{T}^\alpha$ and $\mathcal{T}^\beta$, given a caption string $x = [x_0, ... x_n]$, the tokens after the encoding are $c^\alpha = \mathcal{T}^\alpha(x) = [c^\alpha_0, c^\alpha_1, ....]$, and $c^\beta = \mathcal{T}^\beta(x) = [c^\beta_0, c^\beta_1, ....]$ respectively. Then we construct a mapping $\mathcal{M}_{\alpha \rightarrow \beta}$ from $c^\alpha$ to $c^\beta$ based on their decoding string's overlap. We use $\mathcal{T}^{-1}$ to denote decode function. For example, as in \cref{fig:t_map}, if $\mathcal{T}^{-1}(c_4^\alpha) == \mathcal{T}^{-1}(c_6^\beta)$, then $\mathcal{M}_{\alpha \rightarrow \beta}(4) = 6$; if $\mathcal{T}^{-1}(c_1^\alpha) \in \mathcal{T}^{-1}([c_1^\beta, c_2^\beta, c_3^\beta])$, then $\mathcal{M}_{\alpha \rightarrow \beta}(1) = [1,2,3]$.

Such tokenizer mapping enables the construction of a precise transformation function that aligns the output space of one encoding model to the output space of another. This facilitates seamless interoperability between models employing different tokenizers by allowing consistent translation of encoded representations across diverse tokenization frameworks.

In this work, the VLM we use to calculate confidence score is LLaVA using a text tokenizer LLaMA, and for T2I the text modality the pipeline uses a CLIP tokenizer. We calculate the confidence score $w_k^\beta$ and then use mapping function $\mathcal{M}_{\alpha \rightarrow \beta}$ to calculate the confidence score using 
\begin{equation}
    \mathcal{V} = \mathcal{M}_{\alpha \rightarrow \beta}(k), \quad s_k^\alpha = \frac{1}{|\mathcal{V}|} \sum_{v \in \mathcal{V}} s_v^\beta,
\end{equation}

where \(\mathcal{V}\) denotes the set of elements mapped from index \(k\) under the transformation, \(s_k^\alpha\) and \(s_v^\beta\) represents the $k$-th token score within the model \(\alpha\) and \(\beta\), and \(|\mathcal{V}|\) is the cardinality of the set \(\mathcal{V}\).

\subsubsection{Attention Map Reweighting}
Observing that higher confidence scores tend to correlate with an increased likelihood of a token being associated with hallucinated content, we introduce an inversely proportional weighting coefficient to these scores. Without loss of generality, we employ the following equation as the reweighting mechanism.
\begin{equation} \label{eq:weights}
    w_k = 
\begin{cases}
    - \text{Softmax}(s_k^t) \quad &\text{if}  s_k^t > \epsilon\\
    1 &\text{otherwise}
\end{cases},
\end{equation}
where $\epsilon$ denotes the threshold. We use $s_k^t$ to denote that the score here is calculated using \cref{eq:text_only} and under the model and tokenizer $\alpha$.

In the T2I diffusion framework, usually there are SelfAttention layers and CrossAttention layers. In CrossAttention layer, we use the confidence score to reweight the attention matrix.

\noindent
\textbf{High precision and low recall.}
To determine the appropriate threshold, it is essential that the threshold enables the retrieval of noisy tokens with high precision. Regarding recall, as shown in \cref{fig:hist_d}, the hallucinated content exhibits substantial overlap within the lower range of token scores. Therefore, it is sufficient to prioritize high precision without demanding a similarly high recall rate in this span.

\section{Experiment}

In this section, we begin with a quantitative analysis of hallucination within the T2I training dataset. Then we introduce a carefully designed benchmark InstructBench, which aims to evaluate the model's instruction following ability. Then we report our method's metric on these datasets. 

\subsection{Training Data Construction}
We combine 400K images from Conceputal Captions (CC)~\cite{sharma2018conceptual} and SBU Captions (SBU)~\cite{ordonez2011im2text} to construct our CCSBU dataset for training T2I models. Each image is paired with an visually relevant short description obtained during original data collection. 

We then select two open-sourced captioning models, LLaVA-7B-v1.6 and Share-Captioner, to generate more descriptive and detailed captions for these 400K images.

\subsection{Caption Quality Evaluation} \label{sec:caption_eval}

We evaluate the recaption quality using the following models and evaluation metrics. 

\noindent
\textbf{CLIP} The CLIP-Score is calculated using CLIP model `openai/clip-vit-base-patch32' and the Long-CLIP-Score uses model `BeichenZhang/LongCLIP-L'~\cite{zhang2025long} since it can handle longer contexts. Both CLIP-Score and Long-CLIP-Score are averaged over random selected 10000 images.

\noindent
\textbf{GPT-4o} We adapt GPT-4o for a more fine-grained evaluation. First, we prompt GPT-4o for extracting visible objects from generated detailed captions. For the extracted objects, we further prompt GPT-4o for three times to justify if each object exists in the given image. For each image-caption pair, we count the number of objects as $N$ and calculate the hallucination rate as  \(\frac{\sum_{i=1}^N H_i}{N}\), where \(H_i=\frac{\sum_{j=1}^3\mathbbm{1}(\text{object i not exist})}{3}\).
Detailed prompts are provided in the supplementary.
The average number of objects Num\_Obj and the average hallucination rate Hal\_Rate are calculated over randomly selected 1000 images.
\begin{table*}[th]
\centering

\begin{tabular}{lcccccccc} 
\toprule
Caption Model   & \makecell[c]{Number of \\ Images} & \makecell[c]{Word \\ Count} &  \makecell[c]{CLIP-\\Score (B)} &  \makecell[c]{CLIP-\\Score (L)} & \makecell[c]{LongCLIP-\\Score (B)} &  \makecell[c]{LongCLIP-\\Score (L)} & Num\_Obj & Hal\_Rate \\ \midrule 
LLaVA-7B-v1.6 & 400k  & 118.06     & 33.13   & 29.03   & 31.30     & 24.99          & 6.5       & 16.96   \\
Share-Captioner & 400k & 153.84     & 32.64   & 28.93   & 31.40     & 24.89     & 6.5       & 19.67   \\
Original Text    & 400k      & 10.25      & 31.59   & 27.47   & 32.37    & 28.69        &    2.1       &     18.78 \\\bottomrule  
\end{tabular}
\caption{Evaluation of the recaptioned dataset (CCSBU). Caption model `Original' denotes the original caption. Num\_Obj denotes the number of hallucinated object, and Hal\_Rate denotes the number of hallucinated object over all objects, detailed definination is in \cref{sec:caption_eval}. We calculate all types of CLIP-Score on 10k random CCSBU images. The Num\_Obj and Hal\_Rate are computed on 1k randomly selected CCSBU images. \textbf{All three captions exhibits a relatively high hallucination rate.}} \label{tab:cap_eval}
\end{table*}

\noindent
From \cref{tab:cap_eval}, we derive the following observations:
\begin{enumerate}
    \item Even the most advanced captioning models exhibit significant levels of hallucination, where non-existent objects or attributes are described inaccurately. Notably, models like LLaVA-7B-v1.6 and Share-Captioner show comparable levels of hallucinated object instances and similar hallucination rates, showing persistent challenges across T2I training.
    \item Hallucination rate does not depend on caption length. While captions with limited word counts yield fewer detected objects (Num\_Obj), the proportion of hallucinated objects relative to the total identified objects remains high, indicating that shorter captions do not necessarily improve overall accuracy.
    \item The CLIP-Score may be inadequate as a sole metric for assessing text-image alignment, particularly in cases where noise within captions is both subtle and infrequent. This suggests the need for more nuanced evaluation metrics that can effectively capture and penalize even minor discrepancies in text-image correspondence.
\end{enumerate}

\subsection{InstructBench}

In prior sections, we observed that captions generated by Vision-Language Models (VLMs) frequently exhibit hallucinations and inaccuracies, particularly regarding specific attributes such as object color, spatial relationships among objects, object count, and detailed features. To rigorously assess how these inaccuracies impact the generation capabilities of VLMs across these detailed attribute categories, we developed \textit{InstructBench}. InstructBench comprises four attribute types, each represented by 200 paired generation-evaluation prompts tailored to common hallucination types observed in VLMs, as outlined in \cref{tab:benchmark} and also \cref{fig:visual}. An example is provided for each attribute type, with additional examples available in the supplementary materials.

\noindent
\textbf{Evaluation.}
To assess the congruence of generated images with specified attributes in their captions, we employ a binary "yes" or "no" evaluation prompt directed to GPT-4o. The prompt, exemplified in Table 1, adds a standard instruction suffix, "Answer with 'yes' or 'no'." For each attribute—namely spatial, color, quantity, and feature fidelity—we calculate the proportion of affirmative responses, represented as $A_s$, $A_c$, $A_q$ and $A_f$,  respectively. These ratios signify the accuracy with which the model generates images that align with the specified spatial, color, quantitative, and feature-related attributes. It should be noted that ambiguous or invalid responses, such as "I don't know," are excluded from the affirmative to maintain clarity in evaluation.

\begin{table*}[th]
\centering
\begin{tabular}{llll}
\toprule
Category & Keywords & Generation Prompt & Evaluation Prompt \\ \midrule
{Spatial} &    \makecell[l]{Left, right, vertical, horizontal, \\inside, outside, middle, corner, \etc.}   & \makecell[l]{The dog's head is turned \\ slightly to the right.} &  \makecell[l]{ Is the dog's head turned \\ slightly to the right?}    \\  
{Color} & \makecell[l]{Yellow, orange, purple, pink, \\ brown, light blue, navy blue, \etc.}        &    \makecell[l]{A kitchen with floor of \\ a light blue color.}  &   \makecell[l]{Does the kitchen have a \\ floor of a light blue color?}       \\
{Quantity} & \makecell[l]{Number of objects/living \\ creatures, from one to six.}       &  \makecell[l]{The image shows four \\ individuals posing  \\for a photograph.} &  \makecell[l]{Are there four individuals \\ posing for a photograph?}          \\
{Features} & \makecell[l]{Droopy, locked, unlocked, open, \\closed, full, empty, invisible, \etc} &  A cat with eyes open. & \makecell[l]{Are the cat's eyes open \\ in this image?}        \\ \bottomrule
\end{tabular}
\caption{Category and examples of InstructBench. Each  category comprises 200 generation-evaluation pairs. These categories align with the four primary types of hallucinations typically observed in captioning tasks.}\label{tab:benchmark}
\end{table*}
\subsection{Can The Training Process Effectively Reduce Discrepancies Between Various Captions?}
An inquiry arises regarding the impact of varying captions associated with the same image during the fine-tuning process: \textbf{Does the influence of discrepancies between these captions amplify or diminish over time?} Specifically, if the divergence between the models' outputs decreases throughout fine-tuning, it implies that the differences in captions are negligible and can be effectively disregarded. Conversely, if the models' predictive discrepancies intensify during fine-tuning, this suggests that subtle variations in the captions are not inherently reconciled by the fine-tuning process and may significantly affect the models' convergence.

To explore the aforementioned question, we employed a fixed random seed and fine-tuned the model using datasets with differing captions, resulting in two distinct model checkpoints. We then obtained predictions from both models and calculated the distances between these predictions. The distance metrics were averaged over 192 samples across the nearest twenty iterations, yielding an aggregate average around 4k samples. As illustrated in \cref{fig:dist}, the prediction distances initially increase and then plateau, showing the fact that the discrepancies between these captions will be magnified first then keep constant, thus, showing the signifance to address the noise in caption.

\begin{figure*}[th]
    \begin{center}
    \includegraphics[width=\linewidth]{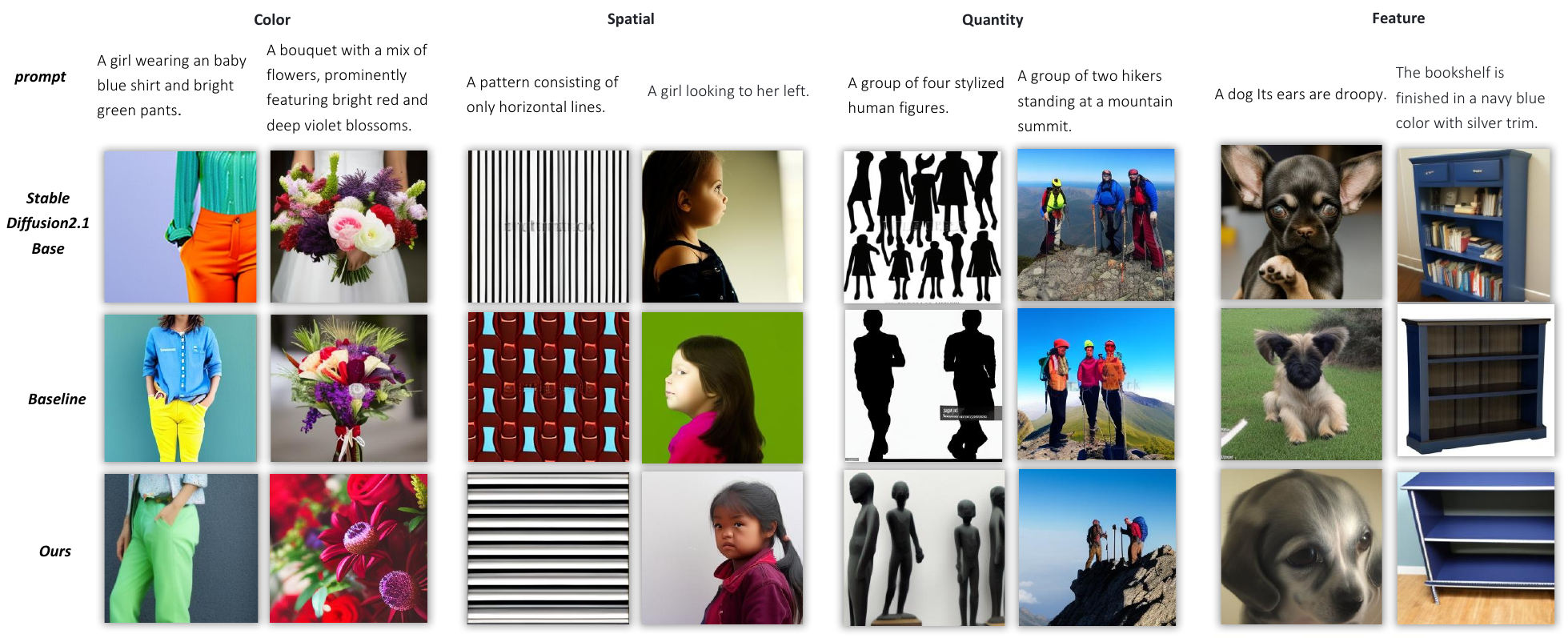}
    \caption{We visualize the generated images and their text prompts across three models on the InstructBench. The top row represents outputs from the original Stable Diffusion 2.1-base model. The middle row is our base model, finetuned on the caption dataset without specific mitigation strategies for caption hallucination. The last row showcases our model trained with the proposed robust training framework. We split the generated images into four dimension: color, spatial, quantity, and feature. We observe that our method better follows the text prompts.}
    \label{fig:visual}
    \end{center}
    \vspace{-0.7cm}
\end{figure*}
\begin{figure}[th]
    \begin{center}
    \includegraphics[width=\linewidth]{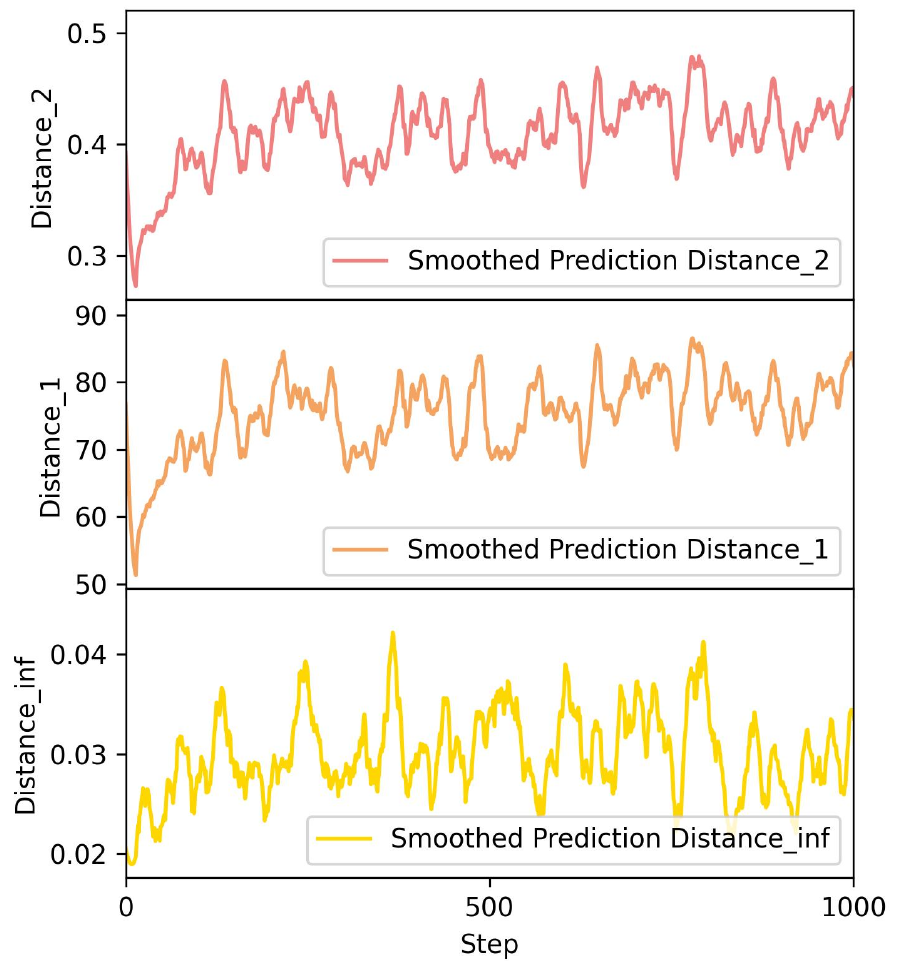}
    \caption{Prediction distance. During finetuning, with a fixed seed, we calculate the $L_1$, $L_2$ and $L_\infty$ distance between the prediction with different captions. All of the three distances first increase and then plateau, indicating that the discrepancies between distinct captions are not reconciled through the fine-tuning process.}
    \label{fig:dist}
    \end{center}
    \vspace{-1cm}
\end{figure}
\begin{table*}[ht]
\centering
\begin{tabular}{lcccccccc}
\toprule
               & CLIP-Score & FID   & $A_s$ (\%) & $A_c$ (\%) & $A_q$ (\%)& $A_f$ (\%) & Ave (\%) & $Acc_l$ (\%) \\ \midrule
Baseline (Original)              & 25.46  & {26.38} & 27 & 43 & 19 & 57 & 36.5 & 82.2      \\ \midrule
Baseline (Share-Captioner)        & 26.03  & 27.13 & 31 & 53 & \textbf{21} & 61 & 41.5 & 89.5      \\
Ours (Share-Captioner)   &  \textbf{26.57}  & \textbf{24.63} &  \textbf{33}    &  55    &  20    &  64    &   \textbf{43.0}  & \textbf{94.7}   \\ \midrule
Baseline (LLaVA-v1.6)                 & 25.94  & 27.25 & 30 & 44 & \textbf{21} & 60 &  38.8 & 85.0 \\
Ours (LLaVA)            &  {26.48}  & 31.50 &  27 & \textbf{56} & 12     & \textbf{66} & 40.3 & 92.5     \\ \bottomrule
\end{tabular}
\caption{Comprehensive evaluation of T2I generation quality. The metrics $A_s$, $A_c$, $A_q$ and $A_f$ quantify the accuracy of correctly generating images with specified captions, focusing on spatial, color, quantity, and features. Ave represents the mean accuracy across these four dimensions. $Acc_l$ assesses the feature representation in middle-layer using linear probing.} 
\label{tab:eval}
\end{table*}

\subsection{Results for Alleviating Caption Noise}
Based on the VLM's confidence score, we propose an approach in \cref{sec:approach} to alleviate the caption noise. We evaluate the proposed method using both general metric (CLIP-Score and FID) and also on the InstructBench for better evaluating its instruction following ability. 

The results are presented in \cref{tab:eval} and supplementary Tab.5, visualization examples in \cref{fig:visual}. The term baseline (original, Share-Captioner, and LLaVA-v1.6), refers to models fine-tuned using the respective captions listed in \cref{tab:cap_eval}. Detailed configurations for these models are provided in the supplementary materials. The ablation `Rm noisy tokens' refers to directly removing tokens with confidence scores exceeding a specific threshold, leaving a subset of tokens that are concatenated to form a refined caption. As shown in \cref{tab:ablation}, we implement several baseline methods for comparison with the proposed approach. The `Rm Noisy Tokens' baseline involves removing tokens identified as hallucinated and concatenating the remaining tokens to generate captions, with hyperparameter $\sigma$=30\%. The `Mix-up' baseline employs the mix-up technique~\cite{zhang2020does,zhang2017mixup} a widely used method for noise robustness, by blending the prediction embeddings and caption embeddings. Additionally, we perform ablation experiments with various hyperparameter settings. the {$\sigma$=10\%} denotes a dynamically calculated threshold corresponding to the lowest 10\% confidence score within a batch, as formulated in \cref{eq:weights}. This thresholding strategy eliminates the need for prior assumptions about caption quality. From these results, we derive several insights.

\begin{enumerate}
    \item For baseline training, the recaptioned dataset demonstrates superior performance compared to the original dataset, suggesting that longer, more detailed captions enhance generation outcomes.
    \item The proposed method incorporating confidence scores, shows consistent improvement across various captions and, more importantly, enhances instruction-following capability, thereby affirming the method's effectiveness.
    \item Although removing noisy tokens directly appears simpler and more straightforward, it leads to a decline in performance compared to the proposed method, indicating that this approach may be overly simplified.
    \item An intermediate value of $\sigma$ results in optimal performance, indicating that when $\sigma$ is too small, most attention map values are influenced, which fails to meet the desired principle of high precision and low recall. Conversely, if $\sigma$ is too large, only a small fraction of tokens are affected, resulting in minimal impact from the method.
\end{enumerate}

\begin{table*}[th]
\centering
\begin{tabular}{cccccccc}
\toprule
                         & CLIP-Score & FID & $A_s$ (\%) & $A_c$ (\%) & $A_q$ (\%)& $A_f$ (\%) & Ave(\%)  \\ \midrule
{Rm Noisy Tokens} & 24.14 & 35.25 & 29 & 44 & 18 & 60 & 37.8 \\
Mix-up \#1 & 25.75 & 25.49 & 30 & 52 & 14 & 56 & 37.5 \ \\
{$\sigma$=10\%} & 25.79 & 28.83 & 28 & 42 & 8 & 55 & 33.3 \\
{$\sigma$=30\%} & 26.57 & 24.63 & 33 & 55 & 20 & 64 & 43.0 \\
{$\sigma$=50\%} &  25.92  & 31.25 &  28 & 56 & 14 & 62 & 40.0 \\
{$\sigma$=70\%} & 26.08 & 28.38 & 29 & 50 & 12 & 58 & 37.3 \\
{$\sigma$=90\%} &  25.64 & 30.75 & 28 & 44 & 10 & 56 & 34.5 \\ \bottomrule
\end{tabular}
\caption{Ablation study. All experiments are using dataset with LLaVA-v1.6 caption. The metrics $A_s$, $A_c$, $A_q$ and $A_f$ quantify the accuracy of correctly generating images with specified captions, focusing on spatial, color, quantity, and features. Ave represents the mean accuracy across these four dimensions.}\label{tab:ablation}
\end{table*}

\subsection{Even Minor Hallucination in Caption Affects The Representation Quality}

To better analyze the caption effect on the T2I generation, we can look into the features to analyze their representation quality~\cite{li2023your}. Especially, following the work~\cite{xiang2023denoising,yu2024representation} which use 
linear probing to analyze the middle-layer representation ability, we adopt the same method in ~\cite{xiang2023denoising} to analyze how the caption quality effect the representation quality. Detailed configuration are in the supplementary. And the results are in \cref{tab:eval} denoted by $Acc_l$ meaning accuracy of linear probing on CIFAR10.

Interestingly, from the results, we can see the different caption can affect the representation quality, and the proposed method can improve the representation quality.

\subsection{Implementation details}

We utilize Stable Diffusion v2.1 base as our backbone for all experiments. We train and inference on 8 NVIDIA A100-SXM4-40GB GPUs. Given the memory constraints, input images are resized to a resolution of $3 \times 256\times 256$. We employ a batch size of 192 (split across devices as 
$24 \times 8$) to optimize GPU utilization while balancing memory efficiency. The learning rate is set at $3e-5$, a common choice for fine-tuning in diffusion-based models to ensure stable convergence. For sampling, we adopt a classifier-free guidance approach with a guidance scale of 7.5 to enhance the fidelity of generated images relative to the input text prompts. The Euler sampler is utilized for efficient generation, with 20 timesteps per sample during training and 1000 timesteps per sample for testing to enable high-quality output at test time. A squared linear schedule is applied to control the noise level during sampling, which helps in maintaining smooth and gradual progression through the denoising steps. To prevent overfitting, we adopt a drop out rate as 0.1.

\section{Limitation and Conclusion}
In this study, we examine the impact of caption noise on text-to-image generation and identify the potential of Vision-Language Model confidence scores as a reliable indicator. We subsequently propose a methodology to leverage these scores through token reweighting to mitigate noise effects. Experiments results demonstrate its effectiveness.

\textbf{Limitations}. While numerous advanced techniques exist for enhancing noise robustness in understanding tasks, their applicability and efficacy in generative tasks remain relatively underexplored. The approach proposed probably is not the most effective or straightforward method. But this paper and findings highlight the critical importance of addressing caption hallucination in T2I systems.

{
    \small
    \bibliographystyle{ieeenat_fullname}
    \bibliography{main}
}

\clearpage
\setcounter{page}{1}
\maketitlesupplementary

\section{Caption Hallucination}
\begin{figure*}[h]
    \begin{center}
    \includegraphics[width=\linewidth]{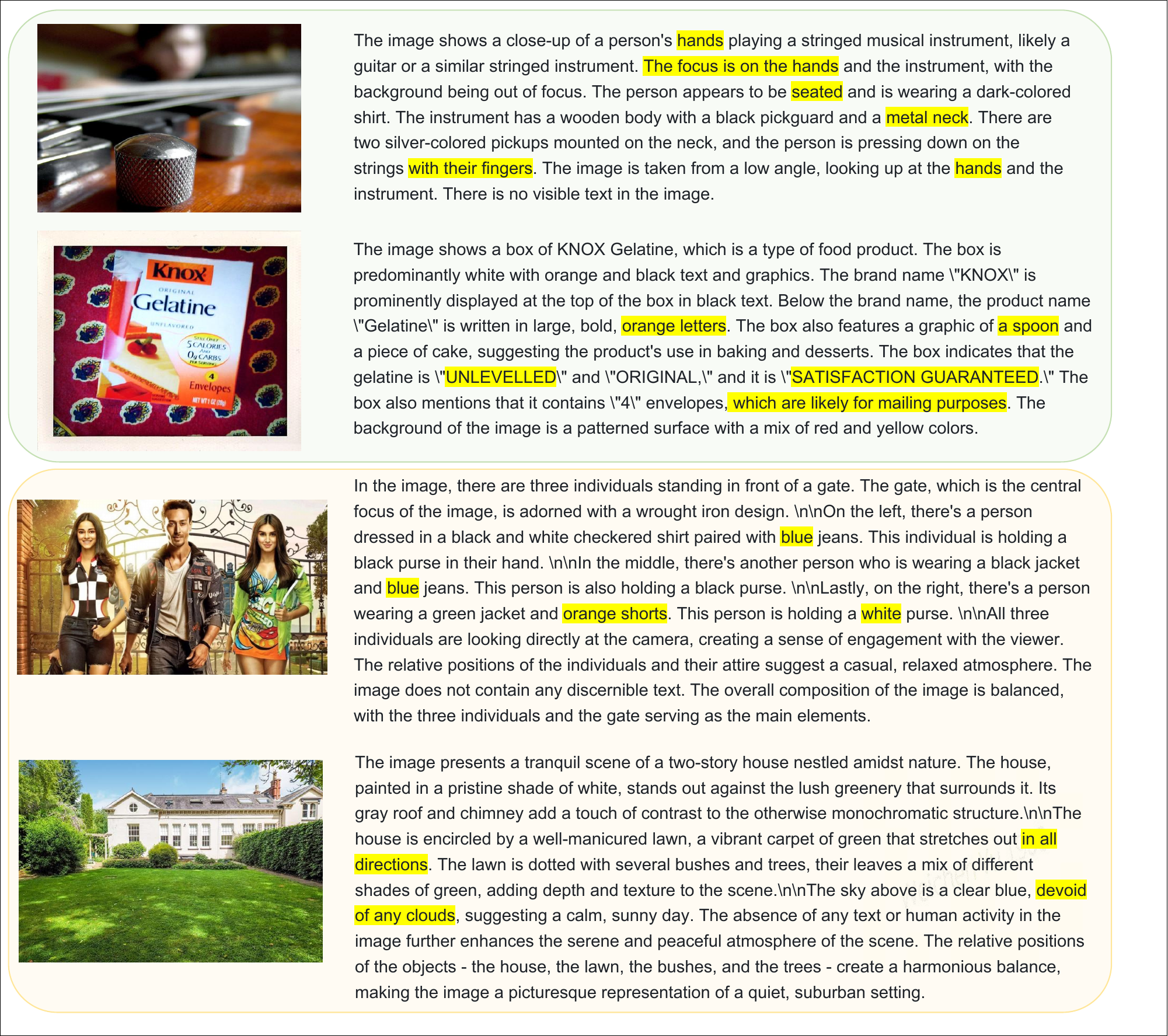}
    \caption{Examples of the caption hallucination in the dataset. The words highlighted in yellow denotes the hallucination which unaligns with the image. The top two rows are captioned using LLaVA-7B-v1.6, and the bottom two rows are captioned using Share-Captioner.}
    \label{fig:cap_noise}
    \end{center}
\end{figure*}

We present examples of hallucinations in captions in \cref{fig:cap_noise}, accompanied by several key observations regarding caption hallucinations:
\begin{enumerate}
    \item \textbf{Variation in Hallucination Types Across Models}: Different captioning models exhibit distinct tendencies in the types of hallucinations they generate. For instance, LLaVA-v1.6 demonstrates a higher prevalence of hallucinations related to color and specific features, while Share-Captioner shows greater susceptibility to hallucinations involving color and spatial relationships.
    \item \textbf{Proportion of Hallucinated Content}: Hallucinated content constitutes only a minor fraction of the overall caption. Current captioning models and vision-language models (VLMs) generally achieve high accuracy, producing predominantly correct captions with hallucinations typically confined to specific attributes.
    \item \textbf{Influence of Text Exposure Bias in Language Models}: Hallucinated content can be attributed, in part, to text exposure bias in large language models (LLMs).\footnote{Text exposure bias refers to the discrepancy between training and testing conditions caused by the reliance on teacher forcing during maximum likelihood estimation.} For example, during LLM pretraining, ``envelopes'' are frequently associated with ``mailing'', which leads to the hallucination of the phrase ``which are likely for mailing purposes'' in the second row of the example.
    \item \textbf{Hallucination Types}: Hallucinations in captions can be categorized into four primary types: color, spatial relationships, quantity, and specific features. To evaluate the quality of generated captions along these dimensions, we have developed a specialized benchmark, InstructBench, designed to assess performance with respect to these attributes.
\end{enumerate}

\section{Prompts for Hallucination Rate}
As described in Section ~\ref{sec:caption_eval}, we prompt GPT-4o for extracting hallucinated contents and calculate the hallucination rate to evaluate the quality of image captions. We design a two-round prompt for objects extraction and hallucination justification separately. Details can be found in Figure~\ref{fig:First_code} and Figure~\ref{fig:Second_code}. 

We first use the prompt in Figure~\ref{fig:First_code} to extract visual components/object from a given detailed image caption. Then, we provide GPT-4o with both the image and the prompt in Figure~\ref{fig:Second_code} to justify the existence of each extracted visual component/object.

\begin{figure*}[h]
\centering
\begin{lstlisting}[language=Python]
# to extract visual components (objects) from a given CAPTION
You are an expert in extracting visual components from image descriptions. 
For the given detailed description, you need to list all the visual components in the description, including objects, 
texture, environment, etc.
Do not count a single object twice. Do not count any conjucture. 
Do not include the atmosphere as visual components. 
Do not include things that are not visible as visual components. 
Do not include motions that have not been done as visual components. 

Here are three samples:

Description: The image shows a spoon is filled with a yellow substance, possibly honey or mustard, and it is being lifted from a bowl. The spoon is held by someone who is not visible in the frame. The background features a wooden table, which adds to the overall homey atmosphere of the scene. The focus is on the spoon and its contents, emphasizing the texture and color of the substance being scooped up.
You should output visual components in this format: ["a spoon", "a yellow substance", "a bowl", "a wooden table"]

Description: The image depicts a street scene with a focus on a building and a car. The building appears to be a two-story structure with a flat roof, possibly a commercial or residential building. There are several cars parked or moving along the street in front of the building. The street itself is lined with trees and a sidewalk, and there are traffic lights visible at the intersection. The sky is partly cloudy, suggesting a fair weather condition. The image is in color and has a standard resolution. There are no visible texts or distinctive brands in the image.
You should output visual components in this format: ["a street scene", "a car", "a two-story structure building with a flat roof", "several cars parked or moving along the street", "the street lined with trees and a sidewalk", "traffic lights at the intersection", "partly cloudy sky"]

Description: The image shows a snake resting in a curved container. The container appears to be made of a material that could be a ceramic or a similar type of enclosure. The snake has a patterned body with shades of brown, black, and yellow. It is coiled up and seems to be in a relaxed state, possibly sleeping or resting. The container is placed on a bed of straw-like material, which provides a naturalistic environment for the snake. The background is not clearly visible due to the close-up nature of the photograph. There are no visible texts or markings on the image.
You should output visual components in this format: ["a snake resting in a curved container", "a ceramic container", "patterned snake body with shades of brown, black, and yellow", "a snake coiled up", "a container placed on a bed of straw-like material"]

Description: {CAPTION}
Can you output the visual components as json following the above format?
\end{lstlisting}

\vspace{-0.1in}
\caption{First prompt used to extract visual components/objects from a given detailed image caption. We provide three in-context-learning examples to instruct GPT-4o for object extraction. The output will follow the json format for parsing.}
\label{fig:First_code}
\vspace{-0.1in}
\end{figure*}

\begin{figure*}[h]
\centering
\begin{lstlisting}[language=Python]
# ask GPT-4o to determine the existence of each extracted object
You are a smart expert in evaluating visual components in images. 
Here is a list of visual components that are possible to exist in the provided image: {}
In the list, visual components are split by the punctuation comma. Please consider them separately.
Ask yourself: Can you see [component] in the image? 
The [component] can be substituted by each element in the list of visual components.
Your answer should be a json of a dictionary of 1/0 answers. 
1 means yes, 0 means no, similar to {{"component_i": 1/0, ...}}
\end{lstlisting}

\vspace{-0.1in}
\caption{Second prompt used to determine the existence of each extracted object from the previous object extraction step. We provide GPT-4o with both input image and the prompt including extracted object to determinine if each extracted object exists in the given image.}
\label{fig:Second_code}
\vspace{-0.1in}
\end{figure*}

\section{Visualization}
\begin{figure*}[th]
    \begin{center}
    \includegraphics[width=\linewidth]{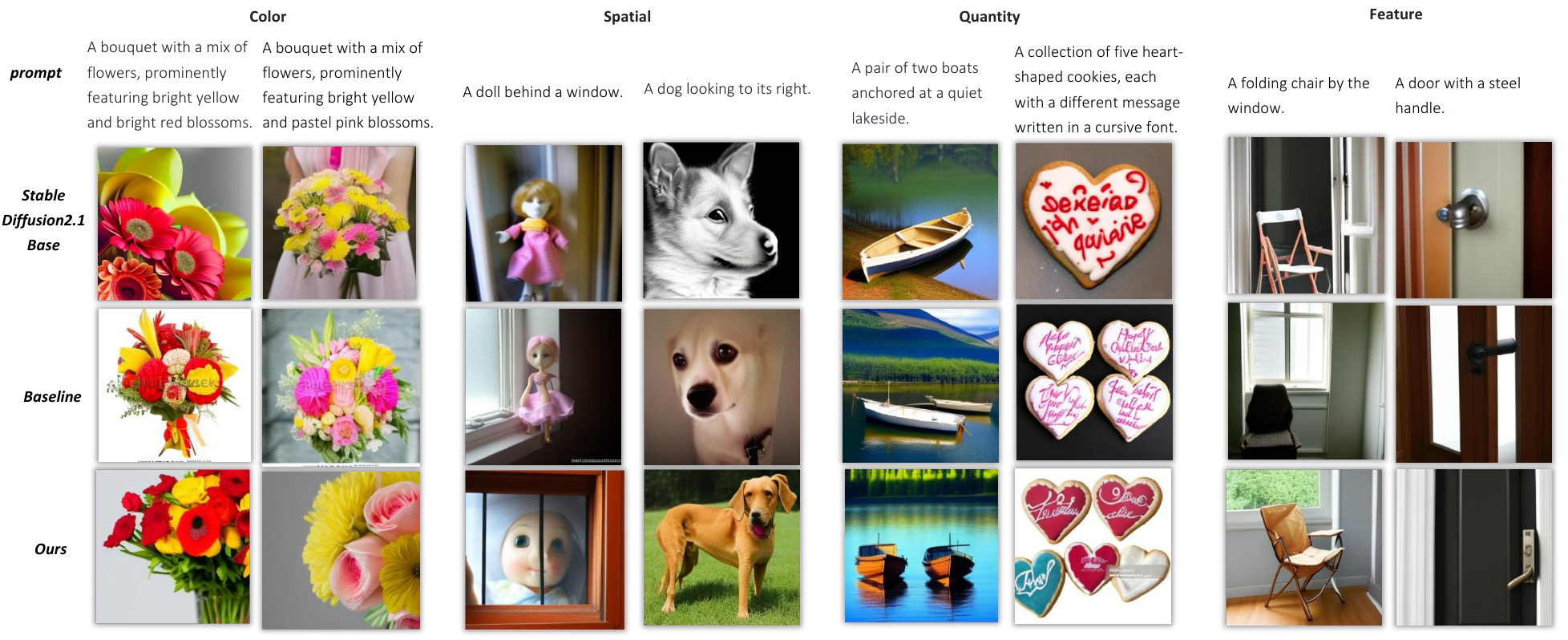}
    \caption{More examples of the generated images on InstructBench. The top row represents outputs from the original Stable Diffusion 2.1-base model. The middle row is our base model, finetuned on the caption dataset without specific mitigation strategies for caption hallucination. The last row showcases our model trained with the proposed robust training framework. We split the generated images into four dimension: color, spatial, quantity, and feature. We observe that our method better follows the text prompts.}
    \label{fig:vis_2}
    \end{center}
\end{figure*}

Apart from Fig.5, we also provide more visualizations in \cref{fig:vis_2}. Here are some observations.

\begin{enumerate}
    \item The proposed method demonstrates a superior ability to adhere to the provided instructions, as evidenced by its higher performance on the InstructBench evaluation benchmark. For instance, in the visualization corresponding to the prompt ``a mix of flowers featuring bright red and bright yellow,'' the proposed method generates images strictly confined to the specified colors of bright red and bright yellow. In contrast, alternative methods may inaccurately include flowers of intermediate hues, such as orange. Similarly, for prompts like ``a folding chair by the window,'' baseline methods frequently fail to satisfy both constraints, either producing an image with a chair that is not folding or omitting the detail of it being positioned ``by the window.''
    \item the proposed method exhibits varying degrees of improvement across different attributes. For attributes such as color fidelity and spatial positioning, the enhancements are particularly pronounced. However, for attributes related to quantity, such as ensuring a specific number of objects, the improvements are comparatively less significant. This suggests that the method excels in some aspects of semantic precision but still faces challenges in others.
\end{enumerate}

\begin{figure*}[th]
    \begin{center}
    \includegraphics[width=\linewidth]{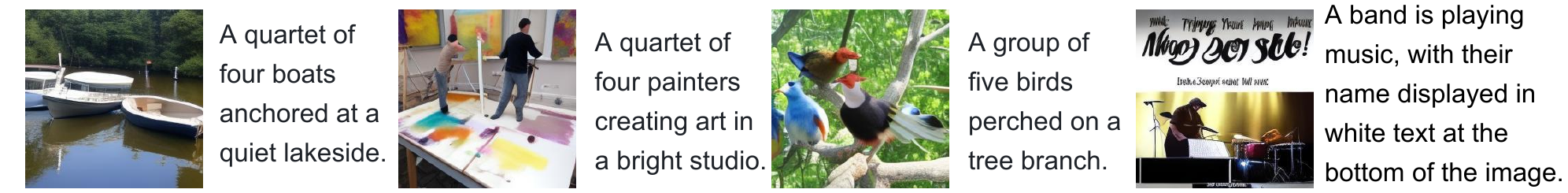}
    \caption{Failure cases of generated images on InstructBench. The failure cases are particularly pronounced in generating quantitative attributes. And a main reason may be the deficiency in training dataset. For instance, in a randomly sampled subset of 10,000 training examples, the term `four' appears 482 times, and `five' appears 89 times, whereas the term `red' occurs 11147 times, and `white' occurs 10619 times. }
    \label{fig:failure_cases}
    \end{center}
\end{figure*}

\noindent
\textbf{Failure Cases Analysis}. We also present several failure cases in \cref{fig:failure_cases}, where the proposed model struggles to generate the intended features. And here are two observations.
\begin{enumerate}
    \item These issues are particularly pronounced in generating quantitative attributes. This limitation likely stems from an imbalance in the training corpus, where quantitative terms appear significantly less frequently than color-related terms. For instance, in a randomly sampled subset of 10,000 training examples, the term `four' appears 482 times, and `five' appears 89 times, whereas the term `red' occurs 11147 times, and `white' occurs 10619 times. 
    \item We observe that the generated images often exhibit watermarks, as at the bottom of the fourth images in the examples, which can confuse the generation model. This artifact arises from the prevalence of watermarked images in the training dataset.
\end{enumerate}

\section{Implementation Details}

In \cref{fig:hist_d}, we use GPT4o-mini to label a subset of hallucinated content, we use the following prompts in the inputs to generate the hallucinated words in the caption.

\begin{mdframed}[backgroundcolor=gray!20]
Hallucinated word generation prompt:

In the following caption of the image, which words are **not** faithfully describing the image?
     List 1)the words and their positions, and a revised version of the caption based on the original caption, in a json format. 
     \{word1: position1, word2: position2,..., caption: revised\_caption\}.
\end{mdframed}

When using VLM to compute the confidence score in \cref{eq:text_only}, we use the following prompts in the inputs of the VLM before the input caption and image.
\begin{mdframed}[backgroundcolor=gray!20]
Prompt for later calculating VLM confidence score:

You are a powerful image captioner. Provide a detailed description of the image.

Instead of describing the imaginary content, only describing the content one can determine confidently from the image. Do not describe the contents by itemizing them in list form. Minimize aesthetic descriptions as much as possible.
\end{mdframed}

\end{document}